\documentclass[sn-nature]{sn-jnl}

\usepackage{graphicx}%
\usepackage{multirow}%
\usepackage{hhline}

\usepackage{graphicx}
\usepackage[11pt]{moresize}
\usepackage{multirow}
\usepackage{array}
\usepackage{caption}
\usepackage{subcaption}
\usepackage{booktabs}%
\usepackage{multicol}%
\usepackage{amsmath,amssymb,amsfonts}%
\usepackage{amsthm}%
\usepackage{subcaption}
\usepackage{mathrsfs}%
\usepackage{lineno}
\usepackage{multirow}
\usepackage[title]{appendix}%
\usepackage{float}
\usepackage{xcolor}%
\usepackage{textcomp}%
\usepackage{manyfoot}%
\usepackage{booktabs}%
\usepackage{algorithm}%
\usepackage{algorithmicx}%
\usepackage{algpseudocode}%
\usepackage{listings}%
\usepackage{graphicx}%
\usepackage{multirow}%
\usepackage{amsmath,amssymb,amsfonts}%
\usepackage{amsthm}%
\usepackage{mathrsfs}%
\usepackage[title]{appendix}%
\usepackage{xcolor}%
\usepackage{textcomp}%
\usepackage{manyfoot}%
\usepackage{booktabs}%
\usepackage{graphicx}
\graphicspath{ {./images/} }
\usepackage{algorithm}%
\usepackage{algorithmicx}%
\usepackage{algpseudocode}%
\usepackage{listings}%
\usepackage{sidecap}

\begin{document}

\title[Article Title]{ESDS: AI-Powered Early Stunting Detection and
Monitoring System using Edited Radius-SMOTE Algorithm}

\author[]{\fnm{A.A. Gde Yogi} \sur{Pramana}}\email{aagdeyogipramana@mail.ugm.ac.id}

\author[]{\fnm{Haidar Muhammad} \sur{Zidan}}\email{haidarmuhammadzidan@mail.ugm.ac.id}

\author[]{\fnm{Muhammad Fazil} \sur{Maulana}}\email{muhammadfazilmaulana@mail.ugm.ac.id}

\author*[ ]{\fnm{Oskar} \sur{Natan}}\email{oskarnatan@ugm.ac.id}

\affil*[]{\orgdiv{Department of Computer Science and Electronics}, \orgname{ Universitas Gadjah Mada}, \orgaddress{\city{Yogyakarta} \postcode{55281}, \country{Indonesia}}}



\abstract{
Stunting detection is a significant issue in Indonesian healthcare, causing lower cognitive function, lower productivity, a weakened immunity, delayed neuro-development, and degenerative diseases. In regions with a high prevalence of stunting and limited welfare resources, identifying children in need of treatment is critical. The diagnostic process often raises challenges, such as the lack of experience in medical workers, incompatible anthropometric equipment, and inefficient medical bureaucracy. To counteract the issues, the use of load cell sensor and ultrasonic sensor can provide suitable anthropometric equipment and streamline the medical bureaucracy for stunting detection. This paper also employs machine learning for stunting detection based on sensor readings. The experiment results show that the sensitivity of the load cell sensor and the ultrasonic sensor is 0.9919 and 0.9986, respectively. Also, the machine learning test results have three classification classes, which are normal, stunted, and stunting with an accuracy rate of 98\%. The official prototype's implementation can be found at \url{https://youtu.be/QOL2TPphINI}.
}


\maketitle
\section{Introduction}\label{sec1}
Stunting is a growth disorder characterized by a shorter height than children of the same age, which impacts cognitive function, productivity, immunity, and neurodevelopment and causes degenerative diseases \cite{WorldHealthOrganization2020Estimates, Titaley2019DeterminantsSurvey}. It is one of the most significant nutritional problems worldwide for children under five. According to WHO, the global stunting rate for children under five is around 22.3\% or 149.2 billion \cite{WorldHealthOrganization2020Estimates}. Additionally, early detection of stunting in children under five is crucial to prevent more severe impacts. However, errors often occur in measuring the height and weight of children under five years due to a lack of skilled personnel and the incompatibility of measuring instruments with anthropometric standards \cite{suparto2022problems, AlAzizah2017PartisipasiSidoarjo}.

A critical challenge in developing machine learning models for stunting detection is dealing with imbalanced datasets. This is where synthetic oversampling techniques like the SMOTE algorithm \cite{Chawla2002SMOTE:Technique} become valuable. SMOTE addresses imbalanced datasets by generating synthetic data points between existing samples and their neighbors. Despite its effectiveness, SMOTE has limitations, such as data overlapping, small disjuncts, noise, and data scarcity  \cite{Saez2015SMOTE-IPF:Filtering}. To overcome these issues, enhanced oversampling techniques, including Borderline SMOTE \cite{Han2005Borderline-SMOTE:Learning} , Safe Level SMOTE \cite{Bunkhumpornpat2009Safe-Level-SMOTE:Problem}, and sample-weighting methods like MWMOTE \cite{Barua2014MWMOTELearning}, ADASYN \cite{2008Neural2008}, and Radius-SMOTE \cite{Pradipta2021Radius-SMOTE:Data}, have been developed.

However, these synthetic oversampling methods are not without drawbacks. They often lack a defined strategy for determining where to generate new synthetic data, risking the creation of data within majority class regions, which can lead to data duplication and other issues. The Radius-SMOTE algorithm \cite{Pradipta2021Radius-SMOTE:Data} addresses these challenges but struggles with small disjunct samples, treating them as noise, which can reduce the classifier's performance. To improve this, the study introduces an enhanced version of Radius-SMOTE specifically designed to better handle small disjunct samples.

To address those issues of stunting detection, it is essential to use advanced solutions for efficient detection and intervention. The use of load cell and ultrasonic sensors offers a promising approach, streamlining stunting detection.  Artificial intelligence can be used to analyse the data from these sensors and machine learning models trained over this data provide insights into what patterns in these indicators of stunting are critical. Based on the aforementioned approaches, the novelty of this research can be listed in the following points.

\begin{itemize}
\item Machine-learning models with inputs related to stunting detection and the use of the Ensemble Machine Learning Model are presented. The model processes and combines multiple algorithms to detect stunting. Also, stunting prediction to clarify their performance based on the confusion matrix is performed.

\item The Ensemble Machine Learning Model is complemented with the novel Edited Radius-SMOTE Algorithm in data Preparation. This allows to apply our model to an imbalanced dataset. Therefore, Edited Radius-SMOTE Algorithm is employed in data preparation stage.
\end{itemize}

\section{Related Works}\label{sec2}
In this section, some related works in the area of stunting detection are reviewed. Then, how they inspire our works is explained.

\subsection{Stunting}\label{subsec1}
Stunting is a linear growth disorder in children due to chronic nutrition, characterized by a z-score height/age $<$ -2 standard deviation as determined by WHO \cite{Titaley2019DeterminantsSurvey}. Stunting is a severe nutritional problem that occurs over a long period due to unmet nutritional needs. Stunting occurs when the fetus is still in the womb and is not visible until the age of two years \cite{Permana2020DETERMINANSTUNTING}. Stunting can be caused by several factors, namely maternal education, economic status, maternal age, maternal number of children, children's health status, and breastfeeding time \cite{Darteh2014CorrelatesGhana}.

In the context of machine learning, these WHO criteria serve as a foundational reference for labeling data during model training. By using these guidelines to classify instances of stunting in the training dataset, the model can learn to predict stunting based on the associated risk factors, rather than relying solely on the z-score threshold. This approach allows for a more nuanced understanding and prediction of stunting risk beyond just the WHO classification, incorporating multiple contributing factors identified through research.

Stunting can be detected early through routine examinations, especially when children are under two years old. The routine examination consists of periodically measuring the child's body mass and length, which is usually carried out at the nearest \textit{posyandu}. Measurements are carried out by trained and skilled health cadres \cite{Syagata2021EvaluasiYogyakarta}.
\subsection{Machine Learning}\label{subsec2}
Machine learning is a branch of artificial intelligence (AI). Machine learning allows computer devices to make decisions in a case through an algorithm that provides the ability to detect and predict patterns. Through a series of algorithms and mathematical models, machine learning can predict data outputs based on pre-existing data \cite{Wang2018MachineOptimization}.

Machine learning models must have as little error as possible so that they can work well when encountering new data. To achieve this, the resulting model must not be similar to the training data because it will not work well when encountering new data. This condition is called overfitting, and the opposite is called underfitting, namely when the model cannot learn differences and patterns in the data \cite{Zhou2021MachineLearning}.
\section{Methodology}\label{sec3}
In this section, the generation of an ESDS hardware prototype the design of an electronic schematic is described. Then, the machine learning modeling and related parameters are explained. The process can be seen in Figure 1 below.
\begin{figure}[h]
\centering
\includegraphics[scale=0.16]{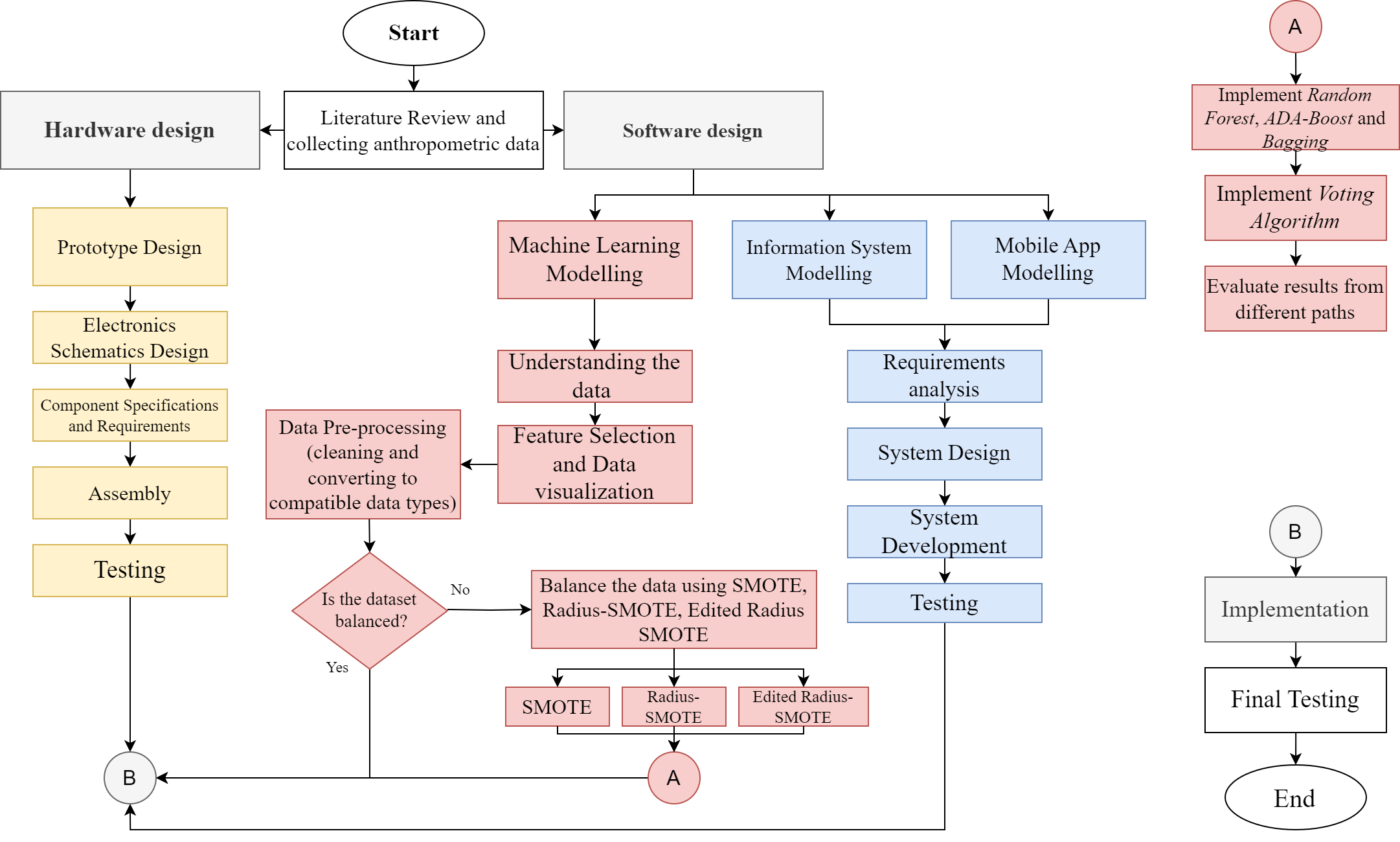}
\caption{Flowchart of Methodology. Yellow blocks represent the flowchart of hardware design. Red blocks are represent for machine learning modeling, while blue blocks are for mobile application modeling in software design.}\label{fig1}
\end{figure}

\subsection{Hardware Design}\label{subsec4}

\subsubsection{Prototype Design}\label{subsubsec1}
\begin{figure}[h]
\centering
\includegraphics[scale=0.068]{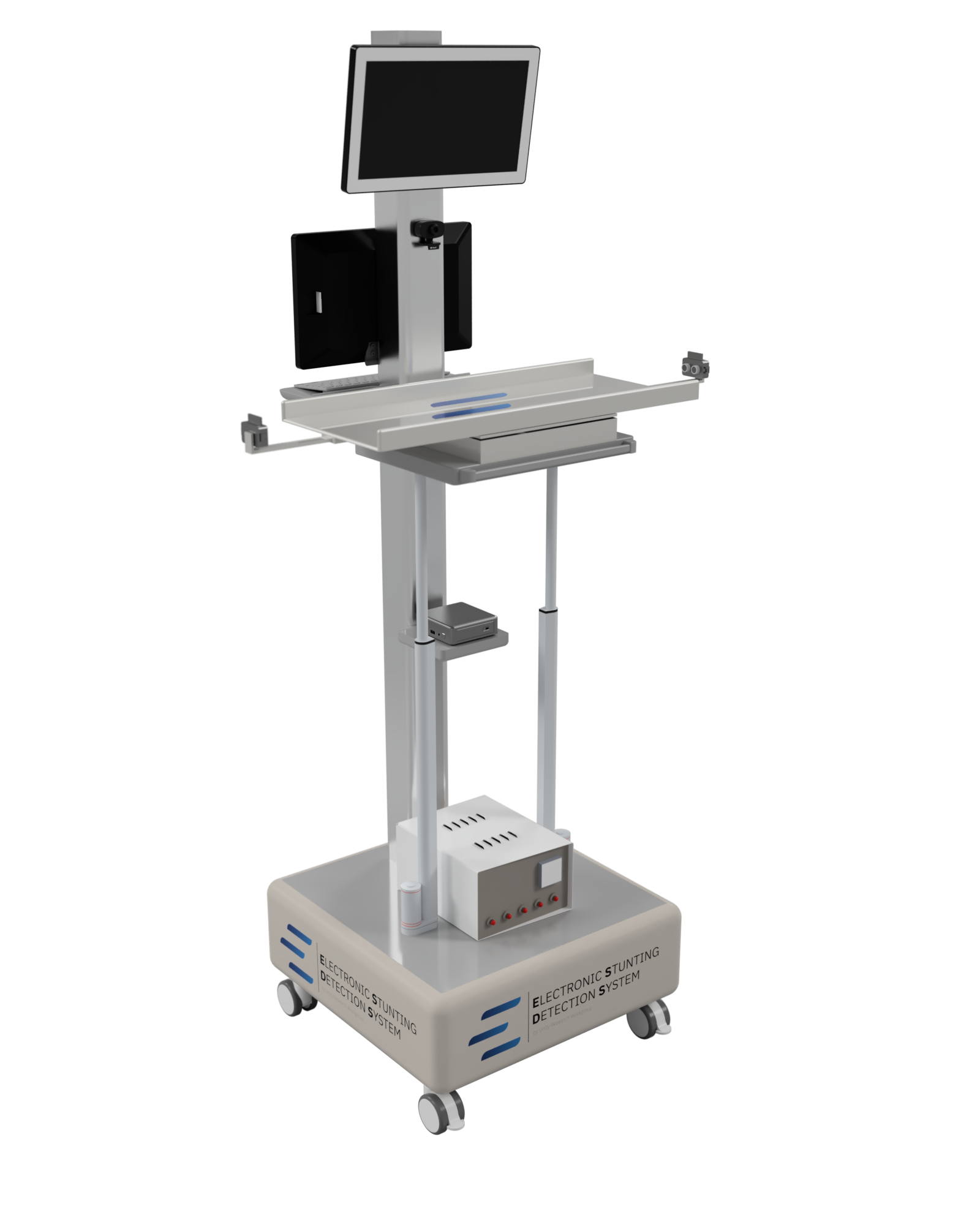}
\caption{The image within the procedure of prototype design process.}\label{fig2}
\end{figure}

The hardware design methodology begins with prototype design phase, where initial concept is created. The hardware design starts from creating a 3D design using Fusion 360 software by considering the placement of components to provide ergonomic value for the user without affecting the accuracy of the antrhopometric equipment. This hardware has dimension of 60 cm x 100 cm x 180 cm which can be seen in Figure 2. 

\subsubsection{Schematic Design}\label{subsubsec2}

\begin{figure}[!h]
\centering
\includegraphics[scale=0.16]{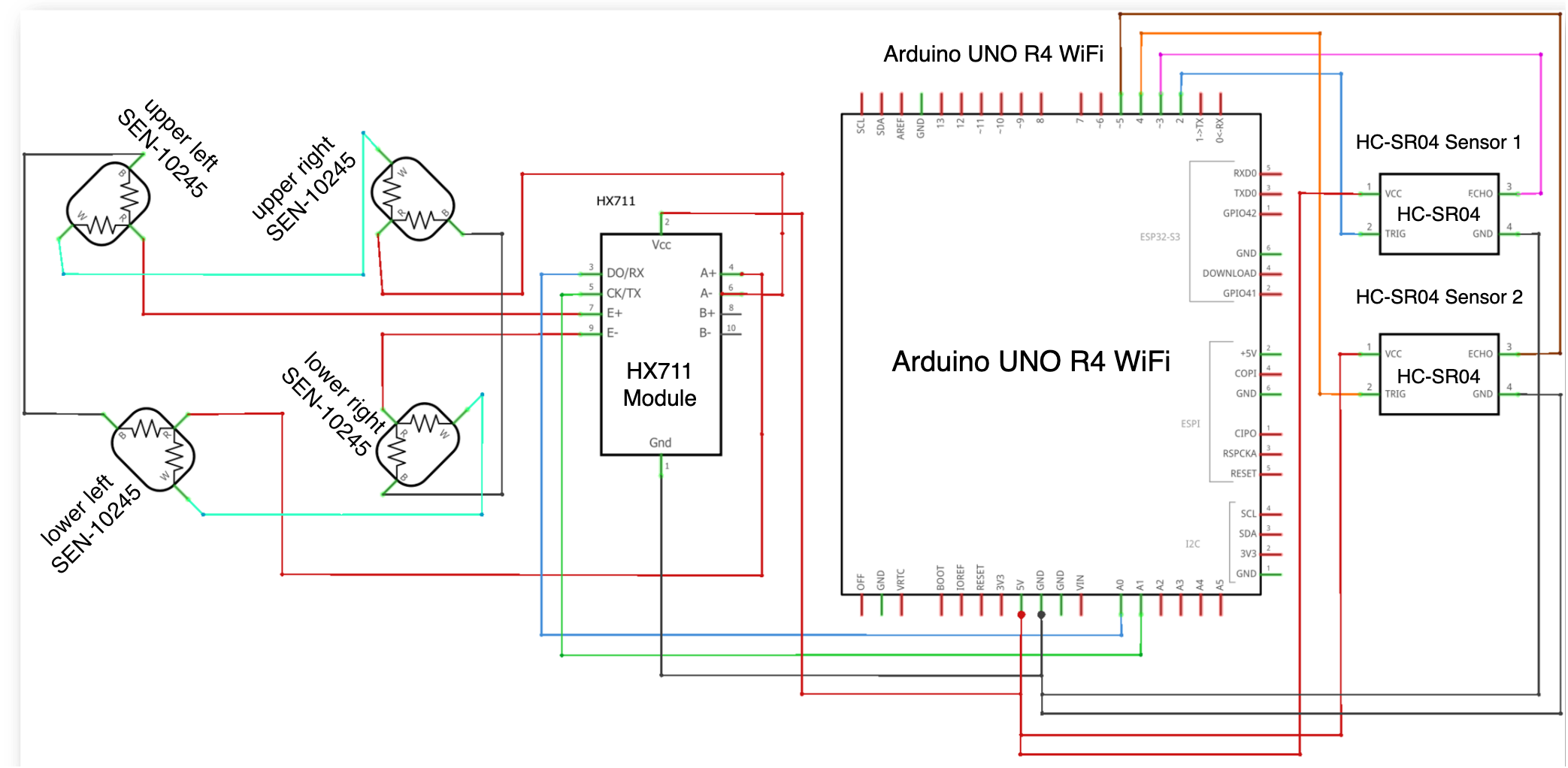}
\caption{The image within the procedure of schematic design process.}\label{fig3}
\end{figure}

Following the prototype design, the electronics schematics design phase involves creating detailed circuit diagrams. This paper uses Fritzing software to create an electronic schematic for the hardware, as shown in Figure 3. The schematics created serve as a roadmap for the assembly process, outlining how each component connects and interacts within the system.

\subsubsection{Component Specification and Requirements}\label{subsubsec3}

\begin{figure}[h]
\centering
\includegraphics[scale=0.18]{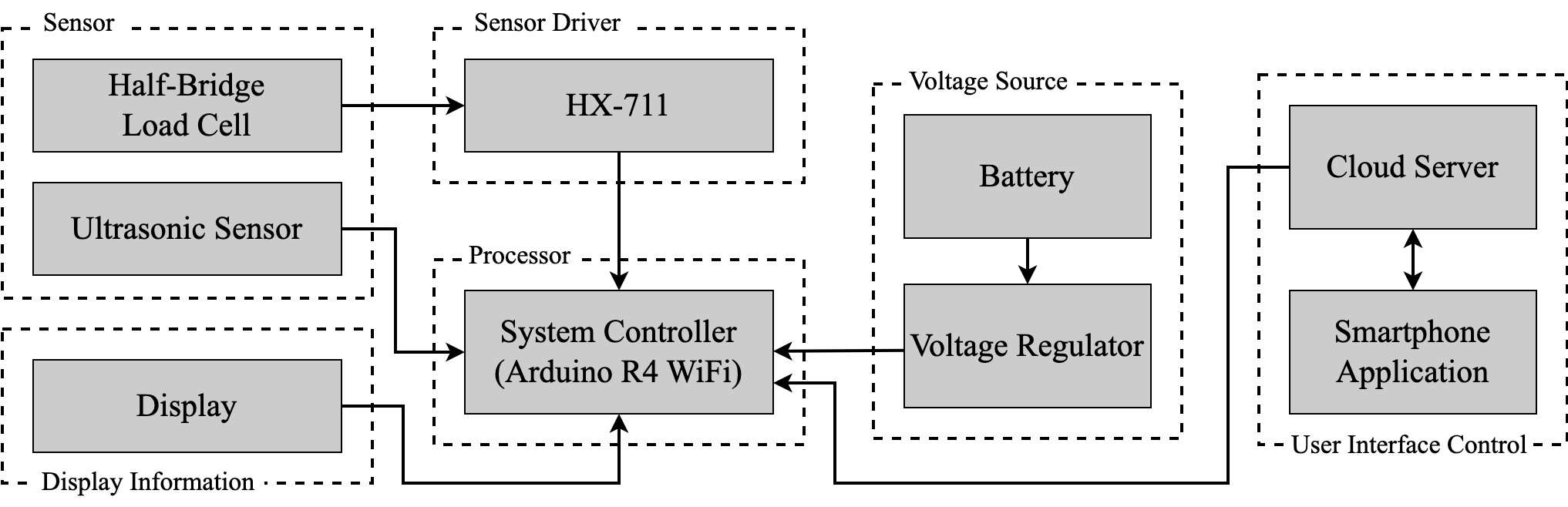}
\caption{Block Diagram of Hardware Component Specification and Requirements}\label{fig4}
\end{figure}

After finalizing the schematic phase, the components and requirements for hardware design are specified and acquired. Based on Figure 4, the half-bridge load cell type is used as the weight sensor, and the HC-SR04 series is used as the height sensor. The load cell used in this paper has a maximum weight of 20 kg. Then, an Analog-to-Digital Converter (ADC) is used to convert the analog output of the load cell sensor into a digital value \cite{Mathew2023DEVELOPMENTMACHINE}, which will be transferred to the microcontroller for data processing. This paper uses HX-711 and Arduino R4 WiFi as the analog-to-digital converter and microcontroller, respectively. Also, the voltage regulator is used to ensures that the voltage supplied from the battery to the microprocessor is stable and within the required range for proper operation \cite{9780380}. After that, the output from the HX-711 and the ultrasonic sensor is processed in the microcontroller. The processed data is displayed in the display information and uploaded to the cloud server so the information is accessible in the smartphone application. Finally, to ensure the accuracy and compatibility of each component, the load cell sensor and the ultrasonic sensor calibration using known weights and lengths are performed, respectively.

\subsubsection{Assembly}\label{subsubsec4}
\begin{figure}[h]
\centering
\includegraphics[scale=0.12]{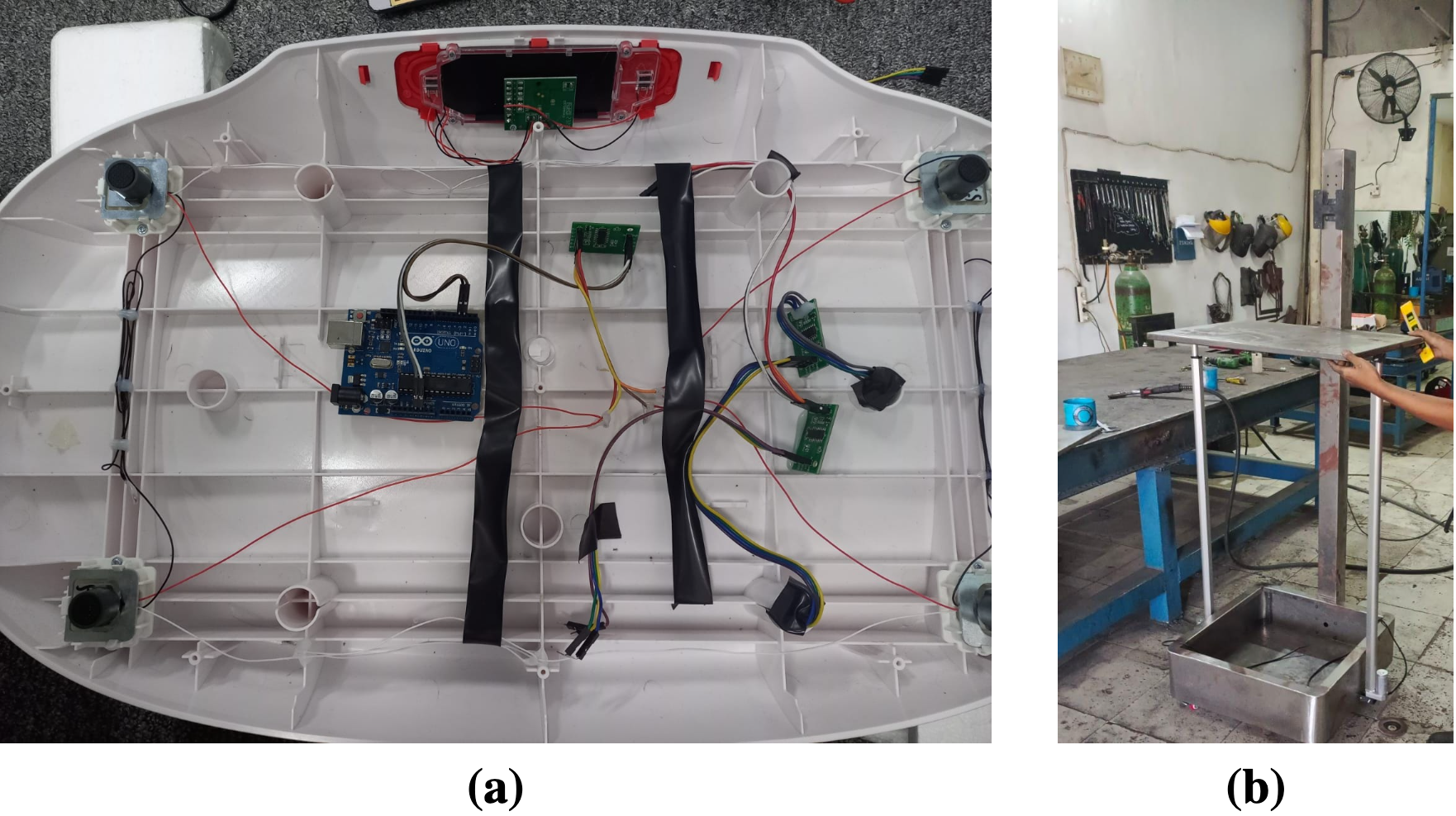}
\caption{The images within the procedure of assembly process. \textbf{(a)} Component assembly. \textbf{(b)} Prototype structure welding process.}\label{fig4}
\end{figure}

The hardware assembly process is divided into three initial stages of making a hardware prototype. The first stage is making schematics of electronic devices and circuits and purchasing the necessary components. The second stage is assembling an initial series of prototypes without a structure and creating a mechanical system that uses CNC and welding processes. After designing the mechanical system and the circuit schematic is complete, proceed to the third stage: combining the mechanical system and electronic circuit into one prototype system. Finally, circuit troubleshooting is carried out to determine and ensure that the problems experienced in the previous stage have been resolved. The assembly and welding process is shown in Figure 5.

\subsection{Software Design}\label{subsec2}
\subsubsection{Machine Learning Model}\label{subsubsec6}
In this research, CRISP-DM (Cross Industry Standard Process for Data Mining) approach is used. This approach consist of 6 phases starting from Business Understanding Phase, Data Understanding Phase, Data Preparation Phase, Modeling Phases, Evaluation Phase, and Deployment Phase \cite{HayatSuhendar2023MachineConcept}.

\paragraph{Business Understanding}
The primary objective of this research is to develop a more accurate and precise machine learning model for stunting detection. Specifically, the goal is to integrate this machine learning model into the Electronic Stunting Detection System (ESDS).

\paragraph{Data Understanding}
To diagnose whether a child is normal, stunted, or severely stunted, we examine how much the child's height/age deviates from the z-score for their age group. For stunted cases, their z-score falls more than 3 standard deviations below the median and for stunting cases, their z-score falls within 3 and 2 standard deviations from the median \cite{MinistryofHealth2020PeraturanAnak}. 

The dataset employs the z-score formula for determining stunting, as defined by the WHO. This research uses data collected from the Yogyakarta region, obtained through collaboration with a local health group in which the organization provided essential data, enabling the production of results using the Machine Learning Model.

\paragraph{Data Preparation}

In Table 1, it can be seen that here are a number of attributes in the dataset that have numerical values instead of categorical data. Originally, the data was recorded as categorical, with the values of gender being ‘male’ and ‘female’ and the values for Status being ‘normal’, ‘stunting’ and stunted’. However, for the model to work optimally the data is encoded and assigned numerical values. Male is assigned as ‘0’ and Female ‘1’. As for Status, normal is ‘0’, stunting is ‘0.5’ and stunted is ‘1’.

\begin{table}[h]
\caption{Table representation of the attributes in our data set}\label{tab2}
\begin{tabular*}{\textwidth}{@{\extracolsep\fill}lccccc}
{} & Age (months) & Gender & Height (cm) & Weight (kg) & Status\\
\midrule
1  & 56 & 0 & 110.0 & 22.7 & 0\\
2  & 33 & 0  & 89.0  & 12.1 & 0\\
...  & ... & ...  & ...  & ... & ...\\
751  & 58 & 1  & 108.8  & 17.1 & 0\\
\botrule
\end{tabular*}
\end{table}

As illustrated in Table 1, our dataset comprises approximately 750 records of children from 
Yogyakarta. The stunting status is indicated in the 'status' column, where 0 signifies 
normal, 0.5 represents stunting, and 1 stunted. According to the status histogram in Figure 
2.6, the data distribution is clearly imbalanced. Table 2 shows the data distribution for the 
status class in more detail. \\

\vspace{-3em}

\noindent
\begin{table}[h!]
\centering
\begin{minipage}{0.4\textwidth}
    \centering
    \captionof{table}{Dataset Data Distribution Before SMOTE}
    \renewcommand{\arraystretch}{0.6}
    \begin{tabular*}{\textwidth}{@{\extracolsep\fill}lcc}
    Status & Count & Percentage\\
    \midrule
    Normal  & 645 & 86 \\
    Stunted  & 89 & 12 \\
    Stunting  & 18 & 2 \\
    \botrule
    \end{tabular*}
\end{minipage}%
\hspace{0.1\textwidth}
\begin{minipage}{0.4\textwidth}
    \centering
    \captionof{table}{Dataset Data Distribution After SMOTE}
    \renewcommand{\arraystretch}{0.6}
    \begin{tabular*}{\textwidth}{@{\extracolsep\fill}lcc}
    Status & Count & Percentage\\
    \midrule
    Normal  & 645 & 34 \\
    Stunted  & 641 & 33 \\
    Stunting  & 623 & 33 \\
    \botrule
    \end{tabular*}
\end{minipage}
\end{table}

Given the imbalanced data condition, it is crucial to apply sampling and data cleaning techniques to enhance the classifier's accuracy. SMOTE (Synthetic Minority Over-sampling Technique) \cite{Chawla2002SMOTE:Technique} is an oversampling method designed to address this issue. The core idea of the SMOTE algorithm is to generate new samples for the minority class. For each minority class sample x, additional samples are randomly selected from its k-nearest neighbors, creating a new sample based on the proximity to these neighbors. This approach, as described in Table I, can lead to the generation of new minority class samples, potentially causing a problem known as sample overlap \cite{Dalianjiaotongdaxue2019ProceedingsChina}.

\begin{equation}
\mathbf{x}_{\text{new}} = \mathbf{x}_{i} + \left| \mathbf{x}_{i}' - \mathbf{x}_{i} \right| \times \partial
\end{equation}

$\mathbf{x}_{\text{new}}$ is the new sample; $\mathbf{x}_{i}$ is the minority sample; $\mathbf{x}_{i}'$ is one of the k-nearest neighbors of $\mathbf{x}_{i}$; $\partial$ is a random number and $\partial \in [0,1]$. \textbf{ENN} Wilson [12] developed the Edited Nearest Neighbor (ENN) algorithm in which $S$ starts out the same as training data sets, and then each instance in $S$ is removed if it does not agree with the majority of its k nearest neighbors (with k=3, typically) [13]. If a sample belongs to the minority class, and there are two or more of its three nearest neighbors that belong to the majority class, then the sample will be removed, thereby leading to smoother boundaries between classes \cite{Dalianjiaotongdaxue2019ProceedingsChina}.

 \textbf{Radius-SMOTE }is a hybrid oversampling technique that combines the strengths of SMOTE oversampling and the built-in KNN classifier [14]. Firstly, the training data are over-sampled by using SMOTE. Secondly, each sample' three nearest neighbors are found in the training data. Thirdly, the samples that are misclassified are removed, producing cleaner data. In this way, not only can we balance the data distribution, but also boundaries between classes are clearer \cite{Dalianjiaotongdaxue2019ProceedingsChina}.

 \textbf{SMOTE – Tomek} is a hybrid sampling method designed for addressing imbalances in datasets. It merges the Synthetic Minority Over-sampling Technique (SMOTE) with Tomek links under-sampling techniques. In the process, it initially employs Tomek links under-sampling to eliminate noisy samples from the majority class. Subsequently, it applies SMOTE to generate synthetic samples for the minority class. This helps to balance the dataset while also reducing the noise in both the majority and minority classes \cite{Kumar2023SMOTE-TOMEK:Prediction}.

\textbf{Random Forests} serve as ensemble learning methods suitable for both classification and regression problems. They consist of numerous decision trees generated from bootstrap samples \cite{Mohandoss2021OutlierClassifier}. \textbf{Bagging} is another ensemble learning method which uses multiple subsets of the training data that are created by randomly sampling with replacement. This means that some data points may be repeated in a subset while others may be omitted \cite{Bbeiman1996BaggingPredictors}. The third ensemble learning method is the \textbf{AdaBoost}. This method adapts and focuses on getting better at the areas where previous models struggled, gradually creating a strong learner from a series of weaker ones \cite{Schapire2003TheOverview}.
\textbf{Voting algorithm} refers to ensemble learning techniques where multiple individual models are combined to make predictions \cite{Kumar2021ClassificationScheme}.

\paragraph{Modeling}
\begin{figure}[!ht]
    \centering
    \includegraphics[width=0.9\linewidth]{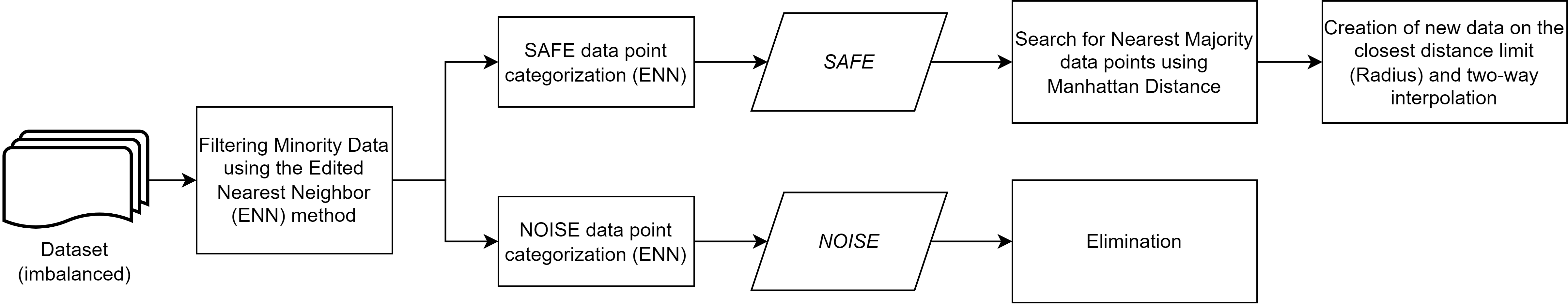}
    \caption{Machine Learning Modeling Flowchart}
    \label{fig:enter-label}
\end{figure}

To detect stunting in our dataset, a robust machine learning model was devised. However, the research began with the essential literature review and data understanding.

According to Figure 6, the research then continued by data pre-processing to check the missing value and doing exploratory data analysis (EDA). After the data is ready, the next step is to process the data with SMOTE, Radius-SMOTE, and Edited Radius-SMOTE Link techniques to balance the data. After the pre-process procedure, we split the data into training (80 percent) and testing (20 percent) types. We put the training data into the ensemble learning classifier namely Ada Boost, Random Forest, and Bagging.

The Voting algorithm is finally implemented in order to combine the different results of the three oversampling techniques.%


\begin{table}[h!]
\centering
\caption{Confusion Matrix for Classification}
\renewcommand{\arraystretch}{0.5}
\begin{tabular*}{\textwidth}{@{\extracolsep\fill}lccc}
\toprule%
& \multicolumn{3}{@{}c@{}}{\ssmall Predicted Label} \\\cmidrule{2-4}%
\ssmall True Label &\ssmall Normal &\ssmall Stunted &\ssmall Stunting \\
\midrule
\ssmall Normal &\ssmall True Negative (TN) &\ssmall False Positive (FP) &\ssmall True Negative (TN) \\
\ssmall Stunted &\ssmall False Negative (FN) &\ssmall True Positive (TP) & False Negative (FN) \\
\ssmall Stunting &\ssmall True Negative (TN) &\ssmall False Positive (FP) &\ssmall True Negative (TN) \\
\botrule
\end{tabular*}
\label{tab:confusion_matrix}
\end{table}

The formulas for calculating this performance measure are given in (2), (3), and (4):






\noindent
\begin{minipage}{.37\linewidth}
\begin{equation}
P(\text{Precision}) = \frac{\text{TP}}{\text{TP} + \text{FP}}
\end{equation}
\end{minipage}%
\begin{minipage}{.34\linewidth}
\begin{equation}
R(\text{Recall}) = \frac{\text{TP}}{\text{TP} + \text{FN}}
\end{equation}
\end{minipage}
\begin{minipage}{.27\linewidth}
\begin{equation}
F1 = \frac{2PR}{P + R}
\end{equation}
\end{minipage}

\paragraph{Evaluation}

For performance measure, precision, recall, and F1-value are taken. Confusion matrix is used and shown in Table 4. The results of the over-sampling algorithm integrated with three ensemble learning algorithm namely Random Forest, Ada Boost, and Bagging as well as the groupings carried out in the testing field.

\paragraph{Deployment}

The proposed methodology leverages various SMOTE algorithm to address class imbalance, enhancing the model's robustness. The integration of ensemble learning further contributes to the overall predictive performance, offering a comprehensive and innovative solution to early stunting detection. 
Our experiment was implemented on MacBook M1 Pro with 16.00G RAM. The results of using the three-over-sampling algorithm SMOTE, Radius-SMOTE, and Edited Radius-SMOTE Link integrated with ensemble learning model Random Forest, Ada Boost, and Bagging. The results of our model can be found in section 4.4. Machine learning model results.

\section{Result and Analysis}\label{sec4}
\begin{figure}[!h]
\centering
\includegraphics[scale=0.125]{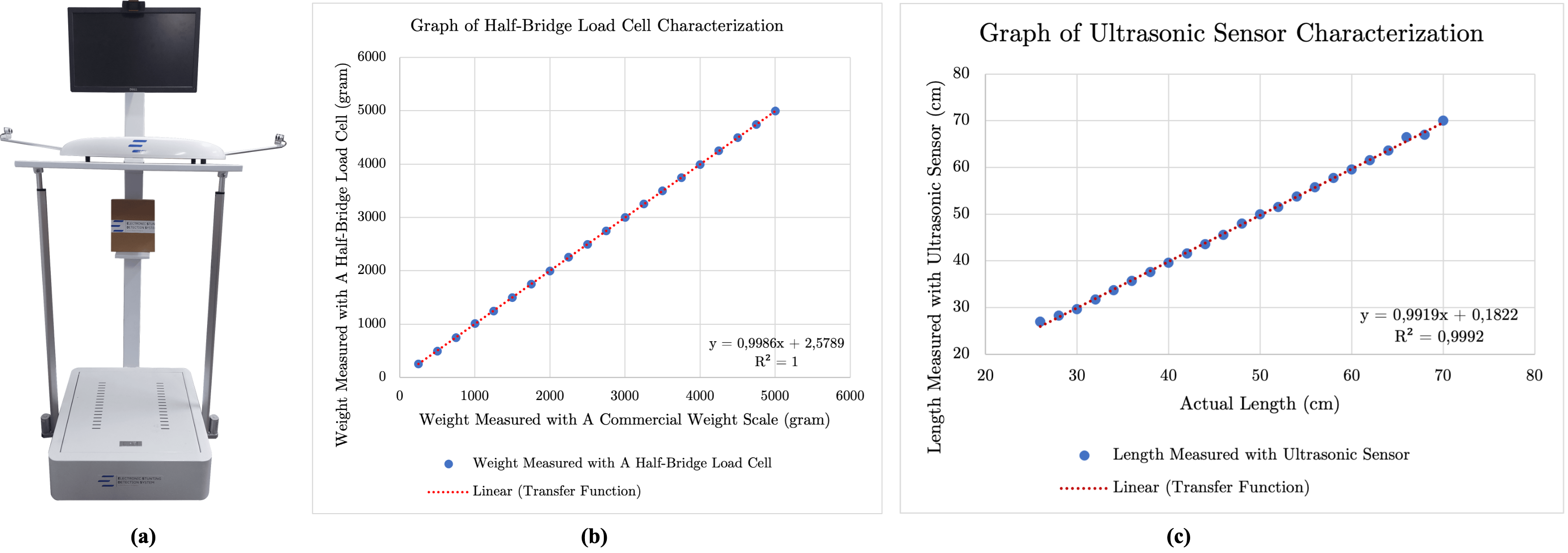}
\caption{The images within the hardware prototype result. \textbf{(a)} Hardware Prototype. \textbf{(b)} The Graph of Half-Bridge Load Cell Characterization. \textbf{(c)} The Graph of Ultrasonic Sensor Characterization}\label{fig4}
\end{figure}

\subsection{Prototype of Developed Anthropometric Equipment}\label{subsubsec7}

The prototype of the developed anthropometric equipment consists of two parts: the structure for anthropometric measuring and the component for the body weight scale and height measuring sensor. The structure has dimension of 60 cm x 100 cm x 180 cm which can be seen in Figure 7(a). 
\subsection{Weight Measurement Components}\label{subsubsec8}
Sensor characterization was performed by comparing the weight measured commercial scale with the output of the half-bridge load cell. Then, the comparison was graphed to obtain the sensitivity value of the load cell sensor, which can be seen in Figure 7(b). The transfer function graphing compares the known weight and the output value obtained. Also, the slope value of 0.9986 means that the load cell sensor has a high sensitivity, so the sensor output is proportional to the known weight value.

\subsection{Height Measurement Components}\label{subsubsec9}


This paper uses two HC-SR04 ultrasonic sensors to measure the length of an object. First, the distances from each sensor (Ultrasonic1 and Ultrasonic2) to the object are measured. The length is then calculated by subtracting these distances from the distance between the two sensors. The transfer function graph in Figure 7(c) shows a comparison between the actual length and the sensor's output, indicating a sensitivity value of 0.9919, meaning the sensor output is nearly proportional to the actual length.

\subsection{Machine Learning Model Results}\label{subsubsec10}

Based on Table 5, we can determine that the random forest algorithm achieves the highest average precision with the SMOTE and Edited Radius-SMOTE methods, both scoring 0.98. The Radius-SMOTE algorithm, on the other hand, has a notably lower precision score of 0.93.

For the AdaBoost algorithm, the Radius-SMOTE method stands out with the highest average precision score of 0.99. It also boasts the highest recall score of 0.99. When it comes to the Bagging algorithm, there is a tie for the highest average precision between the SMOTE and Radius-SMOTE methods, each achieving a score of 0.99. Their recall scores are similarly high, both at 0.99.

Based on these findings, it is evident that our model shows promising results, particularly with the use of the Radius-SMOTE method paired with the Ada Boost Classifier, achieving high precision and recall scores consistently. This indicates that our machine learning model is a satisfactory and reliable choice for the implementation in our anthropometric measurement device, ensuring accurate and dependable performance.

\begin{table*}[!h]
\centering
\caption{Oversampling Result for Random Forest Classifier, Ada Boost Classifier, and Bagging Classifier}
\label{tab:comparison}
\resizebox{1\textwidth}{!}{%
\begin{tabular}{cccccc}
\hline
\textbf{Classifier}         & \textbf{Condition}          & \textbf{Method}             & \textbf{Precision}            & \textbf{Recall}       & \textbf{F-1 Score} \\ \hline
\multirow{9}{*}{\textbf{Random Forest Classifier}} 
                            & \multirow{3}{*}{Normal}     & SMOTE                       & 0.99             & 0.97             & 0.97     \\
                            &                             & Radius-SMOTE                & 0.96             & 0.96             & 0.96     \\
                            &                             & Edited Radius-SMOTE         & 0.97             & 0.98             & 0.98     \\ 
                            & \multirow{3}{*}{Stunted}    & SMOTE                       & 0.95             & 0.98             & 0.97     \\
                            &                             & Radius-SMOTE                & 0.93             & 0.92             & 0.93     \\
                            &                             & Edited Radius-SMOTE         & 0.96             & 0.97             & 0.96     \\ 
                            & \multirow{3}{*}{Stunting}   & SMOTE                       & 1.00             & 0.99             & 1.00     \\
                            &                             & Radius-SMOTE                & 0.98             & 0.98             & 0.98     \\
                            &                             & Edited Radius-SMOTE         & 0.96             & 0.98             & 0.99     \\ \hline
\multirow{9}{*}{\textbf{Ada Boost Classifier}} 
                            & \multirow{3}{*}{Normal}     & SMOTE                       & 0.99             & 0.96             & 0.97     \\
                            &                             & Radius-SMOTE                & 0.99             & 0.99             & 0.99     \\
                            &                             & Edited Radius-SMOTE         & 0.97             & 0.98             & 0.98     \\ 
                            & \multirow{3}{*}{Stunted}    & SMOTE                       & 0.95             & 0.98             & 0.97     \\
                            &                             & Radius-SMOTE                & 0.98             & 0.98             & 0.98     \\
                            &                             & Edited Radius-SMOTE         & 0.96             & 0.97             & 0.98     \\ 
                            & \multirow{3}{*}{Stunting}   & SMOTE                       & 1.00             & 0.99             & 1.00     \\
                            &                             & Radius-SMOTE                & 1.00             & 0.99             & 1.00     \\
                            &                             & Edited Radius-SMOTE         & 1.00             & 0.98             & 0.99     \\ \hline
\multirow{9}{*}{\textbf{Bagging Classifier}} 
                            & \multirow{3}{*}{Normal}     & SMOTE                       & 0.99            &  0.96             & 0.97     \\
                            &                             & Radius-SMOTE                & 0.99            &  0.99             & 0.99     \\
                            &                             & Edited Radius-SMOTE         & 0.97            &  0.98             & 0.98     \\ 
                            & \multirow{3}{*}{Stunted}    & SMOTE                       & 0.95            &  0.98             & 0.97     \\
                            &                             & Radius-SMOTE                & 0.98            &  0.98             & 0.98     \\
                            &                             & Edited Radius-SMOTE         & 0.96            &  0.97             & 0.98     \\ 
                            & \multirow{3}{*}{Stunting}   & SMOTE                       & 1.00            &  0.99             & 1.00     \\
                            &                             & Radius-SMOTE                & 1.00            &  0.99             & 1.00     \\
                            &                             & Edited Radius-SMOTE         & 1.00            &  0.98             & 0.99     \\ \hline
\end{tabular}%
}
\end{table*}

\section{Conclusion and Future Work}\label{sec5}
This paper explores the integration of ultrasonic and half-bridge load cell sensors with machine learning for determining stunting status. The Electronic Stunting Detection System (ESDS) demonstrates that these sensors effectively measure length and weight, with high sensitivity values of 0.9986 and 0.9919, respectively.

The study also employs ensemble learning algorithms, including Random Forest, AdaBoost, and Bagging, to classify data into normal, stunted, and stunting categories. To address data imbalance, oversampling techniques like SMOTE, Radius-SMOTE, and Edited Radius-SMOTE are used. The model’s performance is evaluated using precision, recall, and F1-score, showing that the proposed model better predicts minority classes compared to traditional machine learning methods.

In conclusion, the integration of advanced sensors with ensemble learning algo- rithms offers a promising approach for stunting detection, providing high sensitivity and improved prediction accuracy. Future developments could further enhance this technology, making it more efficient and accessible.

In the future, this research could expand the system to smartphone applications and further enhance sensor technology. There are user interfaces to improve the medical bureaucracy in the field of stunting detection. Then, a study on software development integrated with this system is also interesting study, especially in dealing with machine learning integration into smartphone applications.


\section*{Acknowledgement}
This work was partially supported by the Department of Computer Science and Electronics, Universitas Gadjah Mada under the Publication Funding Year 2024. 

\bibliography{export_z}
\end{document}